\documentclass[conference]{IEEEtran}
\IEEEoverridecommandlockouts
\usepackage{cite}
\usepackage{amsmath,amssymb,amsfonts}
\usepackage{algorithmic}
\usepackage{graphicx}
\usepackage{textcomp}
\usepackage{multirow}
\usepackage{cite}  
\usepackage[table,xcdraw]{xcolor}
\usepackage{scrextend}
\def\BibTeX{{\rm B\kern-.05em{\sc i\kern-.025em b}\kern-.08em
    T\kern-.1667em\lower.7ex\hbox{E}\kern-.125emX}}

\makeatletter
\renewcommand{\footnoterule}{%
  \kern-3pt
  \hrule\@width 1in 
  \@height 0.4pt 
  \kern 2.6pt
}
\makeatother
\begin{document}

\title{PointDico: Contrastive 3D Representation Learning
 Guided by Diffusion Models\\
\thanks{*Corresponding author: sunyiding@stu.xjtu.edu.cn}
}

\author{
\IEEEauthorblockN{1\textsuperscript{st} Pengbo Li}
\IEEEauthorblockA{\textit{International School} \\
\textit{Beijing University of Posts and Telecommunications}\\
Beijing, China \\
2023213418@bupt.cn}
\and
\IEEEauthorblockN{2\textsuperscript{nd} Yiding Sun*}
\IEEEauthorblockA{\textit{School of Software Engineering} \\
\textit{Xi'an Jiaotong University}\\
Xi'an, China \\
sunyiding@stu.xjtu.edu.cn}
\and
\IEEEauthorblockN{3\textsuperscript{rd} Haozhe Cheng}
\IEEEauthorblockA{\textit{School of Software Engineering} \\
\textit{Xi'an Jiaotong University}\\
Xi'an, China \\
chz97@stu.xjtu.edu.cn}
 }

\maketitle

\begin{abstract}
Self-supervised representation learning has shown significant improvement in Natural Language Processing and 2D Computer Vision. However, existing methods face difficulties in representing 3D data because of its unordered and uneven density. Through an in-depth analysis of mainstream contrastive and generative approaches, we find that contrastive models tend to suffer from overfitting, while 3D Mask Autoencoders struggle to handle unordered point clouds. This motivates us to learn 3D representations by sharing the merits of diffusion and contrast models, which is non-trivial due to the pattern difference between the two paradigms. In this paper, we propose \textit{PointDico}, a novel model that seamlessly integrates these methods. \textit{PointDico} learns from both denoising generative modeling and cross-modal contrastive learning through knowledge distillation, where the diffusion model serves as a guide for the contrastive model. We introduce a hierarchical pyramid conditional generator for multi-scale geometric feature extraction and employ a dual-channel design to effectively integrate local and global contextual information. \textit{PointDico} achieves a new state-of-the-art in 3D representation learning, \textit{e.g.}, \textbf{94.32\%} accuracy on ScanObjectNN, \textbf{86.5\%} Inst. mIoU on ShapeNetPart.
\end{abstract}

\begin{IEEEkeywords}
Point Cloud Pretraining, 3D Representation Learning, Diffusion-Contrastive Learning, Multi-modal Knowledge Distillation
\end{IEEEkeywords}

\section{Introduction}
In recent years, self-supervised representation learning has gained significant attention in the fields of computer vision (CV)~\cite{he2022masked} and natural language processing (NLP)~\cite{devlin2018bert}. Pretraining on large-scale datasets provides the model with rich prior knowledge, allowing it to achieve better performance and enhanced generalization capabilities after finetuning, compared to models trained only on downstream tasks. Hence, two prevailing 3D representing methods have each proposed distinct strategies. Generative learning employs pretext task to reconstruct masked point clouds~\cite{pang2022masked}. Contrastive learning, in contrast to generative methods, does not require reconstruction of the original input. Instead, it learns general feature representations by maximizing similarity between positive samples while minimizing similarity between negative samples. 

However, compared to 2D vision and NLP, 3D vision faces a data scarcity challenge. The annotation and collection of 3D data are significantly more time-consuming and costly than those of text and images~\cite{dong2023autoencoders}, resulting in substantially smaller datasets. Meanwhile, contrastive-based methods have been demonstrated to easily identify shortcuts with trivial representations in limited point cloud datasets~\cite{qi2023contrast}. For 3D MAE-based generative methods~\cite{he2022masked}, the unordered nature of point clouds prevents precise one-to-one matching with MSE loss or set-to-set matching with Chamfer Distance loss between the reconstructed and original point, making the reconstruction objective difficult to capture the global density distribution of the object. 

\begin{figure}[t]
\centering
\includegraphics[width=1.0\columnwidth]{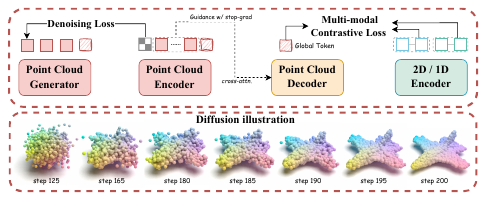}
\caption{\textbf{Schematic illustration of our PointDico.} Our PointDico performs point-to-point denoising from high-noise point clouds to pre-train different backbones. During pre-training, latent features guide multi-level point cloud recovery, while knowledge distillation paradigm is used along with 2D images and language for cross-modal contrastive learning, significantly increasing the diversity of training data. }
\label{ill}
\end{figure}

\begin{figure*}[t]
\centering
\hspace{-0.5cm}
\includegraphics[scale=0.95]{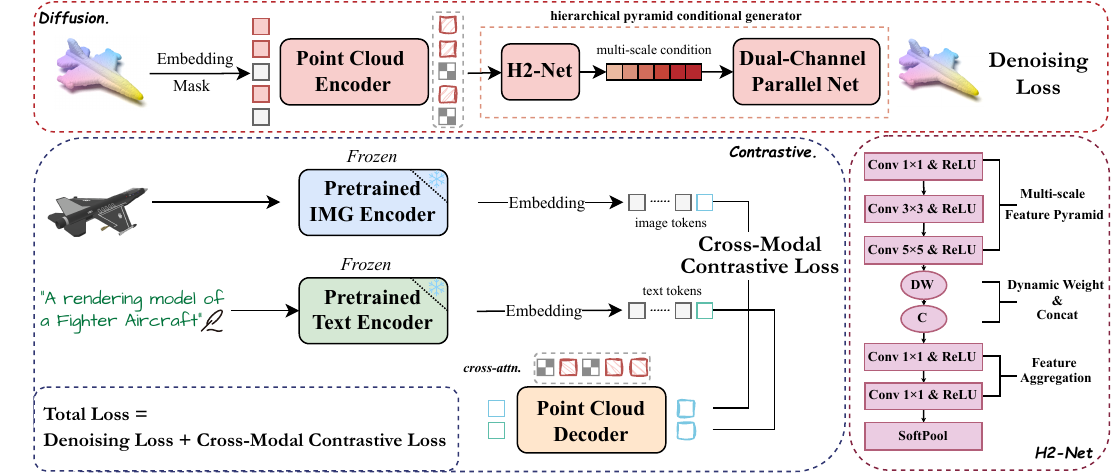}
\caption{\textbf{Pipeline of Our PointDico Framework.} The input modals like point cloud, image, and text are encoded into tokens through respective encoders. For the 3D token embeddings, a hierarchical pyramid conditional generator is applied, combining cross-scale features for denoising. Noted that intermediate embeddings are fed into the 3D decoder via cross-attention with stop-gradient operation. These embeddings are then utilized for cross-modal contrastive learning. }
\label{pipeline}
\end{figure*}

Thus, a critical question arises: \textbf{How can we design a method to overcome their failures to capture global geometric structures in limited and non-uniform point clouds?} 

Inspired by denoising processes of diffusion models, we suppose that Denoising Autoencoders (DAE)~\cite{zheng2024point} offer a superior backbone for point cloud reconstruction compared to MAE. On one hand, by introducing random noise to point clouds, DAE can learn the global structure and distribution characteristics of point clouds instead of merely reconstructing the occluded regions. On the other hand, DAE outperform MAE in handling sparse point clouds and complex geometric structures. In cases where a point cloud exhibits low point density, MAE further diminish the number of visible points through masking, which leads to insufficient information for the reconstruction task. Conversely, DAE preserve the overall point distribution during the noise addition process, thereby maintaining the integrity of the original data.

Considering these advantages, we propose the first diffusion-guided cross-modal contrastive learning paradigm for point cloud representation learning, named PointDico. Specifically, we adopt a hierarchical pyramid conditional generator to guide the point-to-point reconstruction of noisy point clouds. This generator consists of two key components: the Hybrid Hierarchical Network (H2 Net) and the Dual-channel Interaction Point Network (DIP Net). H2 Net incorporates a Multi-scale Feature Pyramid by applying convolution layers of different scales to capture global and local information at various resolutions. A Dynamic Weight Assignment Mechanism further ensures that important features are prioritized by adaptively adjusting the weights of each layer's output through self-attention. DIP Net leverages dual channels to handle high-frequency geometric features (\textit{e.g.}, sharp surfaces or corners) and low-frequency geometric features (\textit{e.g.}, smooth regions or surfaces). The outputs of these two channels are fused through an adaptive gate. Additionally, we employ an encoder-decoder module with cross-attention to transfer semantic information obtained from the diffusion model to cross-modal contrastive learning. From the perspective of knowledge distillation, this process can be regarded as a student network learning invariant knowledge transferred from the encoded representations of the teacher network. By this way, our model leverages multimodal data from 3D point clouds, 2D images, and text for contrastive learning to substantially enhance data diversity.

Our primary contributions can be summarized as follows:
\begin{itemize}
    \item We propose the first diffusion-guided contrastive learning framework for point cloud pre-training, named \textit{PointDico}. By iteratively denoising noisy point clouds and transferring semantic information to cross-modal contrastive learning through knowledge distillation, we achieve a more comprehensive geometric prior.
    \item We introduce a hierarchical pyramid conditional generator composed of H2 Net and DIP Net. This generator facilitates the capture of geometric features at various resolutions while closely integrating global and local point cloud features through cross-scale feature interaction.
    \item Our \textit{PointDico} demonstrates competitive performance across various real-world downstream tasks. Through extensive experiments, we show that the diffusion-contrastive paradigm outperforms both MAE and DAE settings.
\end{itemize}

\section{Related Works}
\textbf{Contrastive Representation Learning.} Contrastive representation learning aims to extract meaningful feature representations by maximizing the similarity between positive pairs while minimizing it with negative pairs. Early works emphasized various strategies to improve representation quality. SimCLR~\cite{chen2020simple} stressed the importance of strong data augmentations, projection heads, and large batch sizes. BYOL~\cite{grill2020bootstraplatentnewapproach} eliminated the need for negative samples by adopting an online-target network, where the target network was updated using momentum, thus improving simplicity and scalability.

In the 3D domain, PointContrast~\cite{xie2020pointcontrastunsupervisedpretraining3d} adapted contrastive learning to unordered point cloud data by contrasting features across augmented views, enabling robust generalization for 3D segmentation tasks. To address data sparsity and modality integration, RECON~\cite{qi2023contrast} combined contrastive and generative approaches, balancing global and local feature extraction with hybrid objectives. Unlike previous methods, our PointDico pioneers the use of diffusion models to guide cross-modal contrastive learning, enriching point cloud data and significantly enhancing generalization performance.

\textbf{Masked autoencoders.} 
Masked autoencoders (MAE)~\cite{he2022masked} have significantly advanced self-supervised learning for 3D point cloud analysis, with numerous extensions refining their design. Point-MAE~\cite{pang2022masked} introduced a transformer-based encoder-decoder framework aimed at reconstructing masked point patches to capture global structures. Later, hierarchical strategies like Point-M2AE~\cite{zhang2022point} improved feature modeling by employing pyramid masking, enabling better representation of fine-grained details and global geometries. ACT~\cite{dong2023autoencoders} incorporated 2D image features to improve cross-modal pretraining. PCP-MAE~\cite{zhang2024pcpmaelearningpredictcenters}, reduced reliance on positional embeddings by integrating patch center prediction, thereby enriching semantic representations. Dual-branch architectures, exemplified by Point-FEMAE~\cite{zha2023towards}, combined global context and local detail extraction using Local Enhancement Modules (LEMs).

In contrast to MAE-based methods, \textit{PointDico} achieves more effective learning of global geometric structures and distribution patterns. 3D MAE rely heavily on reconstructing masked patches, which limits their ability to handle noise and missing data. \textit{PointDico}, on the other hand, leverages a noise addition process that acts as a regularization mechanism to improve generalization. By integrating 3D diffusion generation, \textit{PointDico} enhances robustness against incomplete or noisy data and produces more accurate representations.

\textbf{Denoising diffusion models.} Denoising diffusion models have become a groundbreaking framework in image generation~\cite{pmlr-v37-sohl-dickstein15}. Utilizing a forward Gaussian diffusion process and a backward generation process, denoising diffusion models iteratively refine the generated image starting from Gaussian noise. This process has proven extremely powerful for rich text-conditioned generation of both images~\cite{Ramesh2022HierarchicalTI}, point clouds~\cite{luo2021diffusion} and videos~\cite{10.5555/3600270.3600898}.

However, few works attempted to explore the full potential of diffusion models to pretrain backbones by recovering the original state from random noise. DiffMAE~\cite{wei2023diffusion} pretrained the 2D vision network by denoising pixel values of masked patches. PointDif\cite{zheng2024point} extended diffusion model in 3D vision by presenting a conditional point generator to guide the
point-to-point generation from the noisy point cloud. To our knowledge, we are the first to apply diffusion models to guide 3D cross-modal contrastive learning. Our \textit{PointDico} utilizes knowledge distillation to transfer semantic information captured by diffusion models into a cross-modal contrastive learning framework, effectively addressing the data scarcity challenge in point cloud processing.

\section{METHOD}
\subsection{Conditional Point Diffusion}
In the diffusion framework, noise is iteratively introduced to a point cloud following a Markov chain, establishing a mapping relationship between noisy point clouds across successive steps. Formally, a clean point cloud $P^0 \in \mathbb{R}^{n \times 3}$, consisting of $n$ points sampled from the distribution $p_{\text {data}}$, is corrupted by Gaussian noise as follows:
\begin{equation}
q\left(P^{1: T} \mid P^0\right)=\prod_{t=1}^T q\left(P^t \mid P^{t-1}\right),
\end{equation}
\begin{equation}
q\left(P^t \mid P^{t-1}\right)=\mathcal{N}\left(P^t ; \sqrt{1-b_t} P^{t-1}, b_t I\right),
\end{equation}
where $b_t$ represents predefined constants that gradually increase over time. At time step $t, P^t$ is sampled from a Gaussian distribution with mean $\sqrt{1-b_t} P^{t-1}$ and variance $b_t I$. Additionally, the noisy point cloud $P^T$ at the final step $T$ can be expressed directly in terms of $P^0$ as:
\begin{equation}
q\left(P^T \mid P^0\right)=\mathcal{N}\left(P^T ; \sqrt{\bar{a}_T} P^0,\left(1-\bar{a}_T\right) I\right),
\end{equation}
where $\bar{a}_T=\prod_{i=1}^T a_i$ with $a_T=1-b_T$. As $t$ approaches $T, \bar{a}_T$ asymptotically approaches zero, and $q\left(P^T \mid P^0\right)$ becomes indistinguishable from Gaussian noise $p_{\text {noise }}$.

\begin{figure}[t]
\centering
\includegraphics[width=1.0\columnwidth]{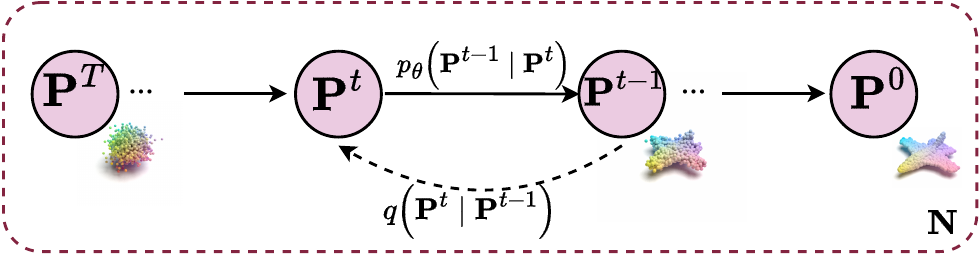}
\caption{\textbf{The directed graphical model of the diffusion process for point clouds.} $\mathcal{N}$ is the number of points in the point cloud $P^0$. }
\label{cond}
\end{figure}

In the reverse process, as shown in Fig. \ref{cond}, Gaussian noise is iteratively removed from the point cloud under the guidance of a condition $c$, using a neural network parameterized by $\theta$. The denoising process is defined as:
\begin{equation}
p_\theta\left(P^{0: T}, c\right)=p\left(P^T\right) \prod_{t=1}^T p_\theta\left(P^{t-1} \mid P^t, c\right),
\end{equation}
where 
\begin{equation}
p_\theta\left(P^{t-1} \mid P^t, c\right)=\mathcal{N}\left(P^{t-1} ; \mu_\theta\left(P^t, t, c\right), \sigma_t^2 I\right).
\end{equation} 

Here, $\mu_\theta$ predicts the mean using a neural network, and $\sigma_t^2$ is a time-dependent constant. The posterior distribution for the reverse diffusion process $q\left(P^{t-1} \mid P^t, P^0\right)$ depends on the initial point cloud $P^0$ and can be expressed as:
\begin{equation}
q\left(P^{t-1} \mid P^t, P^0\right)=\mathcal{N}\left(P^{t-1} ; \tilde{\mu}_t\left(P^t, P^0\right), \tilde{b}_t I\right),
\end{equation}
where the mean $\tilde{\mu}_t\left(P^t, P^0\right)$ and variance $\tilde{b}_t$ are defined as:
\begin{equation}
\tilde{\mu}_t\left(P^t, P^0\right)=\frac{1}{\sqrt{a_t}}\left(P^t-\frac{b_t}{\sqrt{1-\bar{a}_t}} \epsilon\right),
\end{equation}
\begin{equation}
\tilde{b}_t=b_t \frac{1-\bar{a}_{t-1}}{1-\bar{a}_t}.
\end{equation}

The training objective of the diffusion model is derived from variational inference, utilizing the variational lower bound ($v l b$) to optimize the negative log-likelihood:
\begin{equation}
\begin{aligned}
\mathcal{L}_{v l b} = & \mathbb{E}_q\Big[ -\log p_\theta\left(P^0 \mid P^1, c\right) \Big. \\
& + D_{\mathrm{KL}}\left(q\left(P^T \mid P^0\right) \| p\left(P^T\right)\right) \\
& \Big. + \sum_{t=2}^T D_{\mathrm{KL}}\left(q\left(P^{t-1} \mid P^t, P^0\right) \| p_\theta\left(P^{t-1} \mid P^t, c\right)\right)\Big],
\end{aligned}
\end{equation}
where $D_{\mathrm{KL}}(\cdot)$ is the KL divergence. Following PointDif~\cite{zheng2024point}, we adopt a simplified version of the mean squared error:
\begin{equation}
\mathcal{L}(\theta)=\mathbb{E}_{t, P^0, c, \epsilon}\left[\left\|\epsilon-\epsilon_\theta\left(\sqrt{\bar{a}_t} P^0+\sqrt{1-\bar{a}_t} \epsilon, c, t\right)\right\|^2\right],
\label{loss}
\end{equation}
where $\epsilon \sim \mathcal{N}(0, I), \epsilon_\theta(\cdot)$ is a trainable neural network that takes the noisy point cloud $P^t$ at time $t$, along with the time $t$ and condition $c$ as inputs. This network predicts the added noise $\epsilon$. 

\subsection{Hierarchical Pyramid Conditional
Generator}
Our Hierarchical Pyramid Conditional Generator primarily consists of two components: \textbf{H2 Net} and \textbf{DIP Net}.

\textbf{Hybrid Hierarchical Network (H2 Net):} In H2 Net, we integrate a multi-scale feature pyramid to capture both global and local information across multiple resolutions. Specifically, the visible patch features $\left\{T_i^v\right\}_{i=1}^g$ and masked patch features $\left\{T_i^m\right\}_{i=1}^r$, extracted by the encoder (see \textit{Section \ref{encoder}}), are concatenated while preserving their spatial positional information. The concatenated features are then processed through three convolutional layers with kernel sizes $1 \times 1,3 \times 3,5 \times 5$ to extract features at different receptive field sizes:
\begin{equation}
F_{m s}=\left\{f^{1 \times 1}(T), f^{3 \times 3}(T), f^{5 \times 5}(T)\right\},
\end{equation}
where $F_{m s}$ represents the set of multi-scale features produced by the convolutional layers. In this way, both fine-grained local details (captured by $1 \times 1$ convolutions) and broader contextual information (captured by $3 \times 3$ and $5 \times 5$ convolutions) are effectively encoded.

To prioritize critical features across multi-scale outputs, we introduce a dynamic weight assignment mechanism based on attention. 
Let the weight for the $k$-th scale be $\alpha_k$, calculated as:
\begin{equation}
\alpha_k=\frac{\exp \left(h_k\right)}{\sum_{j=1}^K \exp \left(h_j\right)}, \quad h_k=\operatorname{MLP}\left(F_k\right),
\end{equation}
where $h_k$ represents the raw attention score for scale $k$, and $K$ is the number of scales. The multiscale features are then adaptively combined as:
\begin{equation}
F_{\text {combined }}=\sum_{k=1}^K \alpha_k \cdot F_k .
\end{equation}

The aggregated feature $F_{\text {combined }}$ is passed through a series of $1 \times 1$ convolutional layers to reduce dimensionality and produce the guiding hierarchical pyramid condition $c$, which is required for the conditional point cloud diffusion model:
\begin{equation}
c=f_\omega\left(F_{\text {combined }}\right),
\end{equation}
where $f_\omega(\cdot)$ denotes the sequence of $1 \times 1$ convolutional and soft-pooling layers with the parameter $\omega$.

\begin{figure}[t]
\centering
\includegraphics[width=1.0\columnwidth]{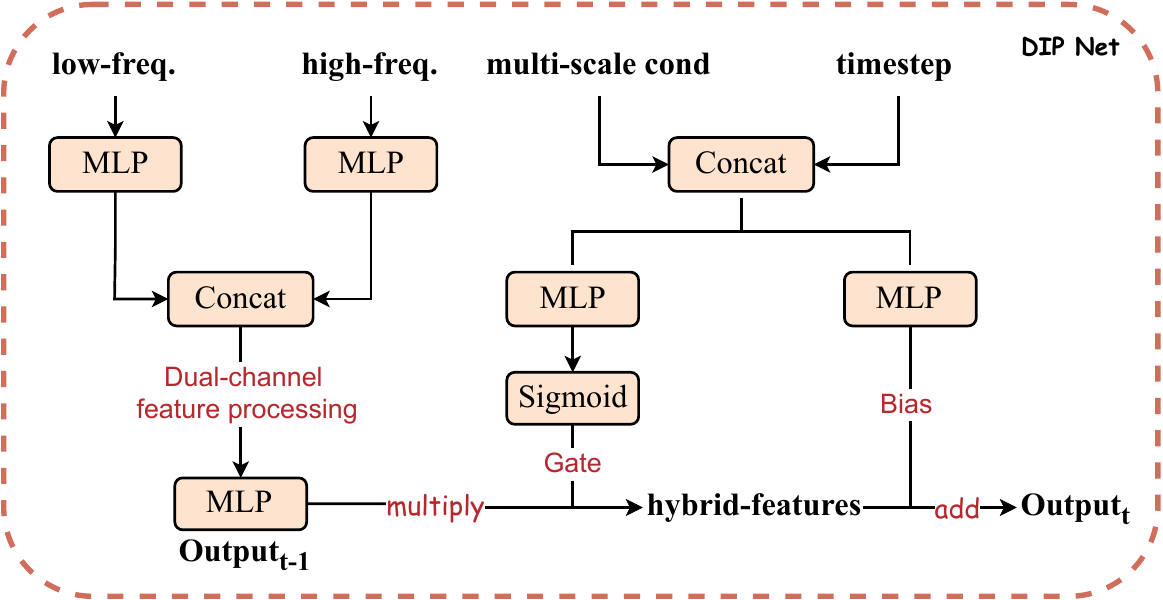}
\caption{\textbf{The illustration of the proposed mode.} By leveraging a dual-channel architecture and a cross-scale interaction mechanism, DIP Net enables an efficient integration of local and global contextual information, while maintaining the flexibility to adapt its feature representation based on external conditions.}
\label{dip}
\end{figure}

\textbf{Dual-channel Interaction Point Network (DIP Net):} In order to address challenges in multi-scale feature extraction and adaptive condition processing for dynamic point cloud representations, we adopt a DIP Net by integrating local and global contextual information through a dual-channel design in Fig.\ref{dip}. The high-frequency branch extracts fine-grained local details, such as sharp edges or complex geometric structures, while the low-frequency branch captures global structural information, such as smooth regions and overall shapes. These features are concatenated along the channel dimension to form a unified representation:
\begin{equation}
F_{\text {concat }}=\operatorname{Concat}\left(F_{\text {high }}, F_{\text {low }}\right) .
\end{equation}

To recover a randomly perturbed point cloud in a point-wise way, DIP Net employs a conditional modulation. External conditions, represented as a combination of timestep embedding  and contextual condition embedding ( $B \times 1 \times C$ ), are processed through MLP layers. A gating vector $G$ with the sigmoid activation, and a bias vector $B$ are used to modulate the features:
\begin{equation}
\begin{split}
&G=\operatorname{Sigmoid}(\operatorname{MLP}(\operatorname{Concat}(\text {$c$ , TE}))), \\
& B=\operatorname{MLP}(\operatorname{Concat}(\text {$c$ , TE})), \\
&F_{t}=\operatorname{MLP}(F_{t-1}) \cdot G+B,
\end{split}
\end{equation}
where TE denotes the timestep embedding.

DIP Net is iterated six times, with input dimensions for each iteration being $[3,128,256,512,256,128]$. The final output dimension is 3. By combining hierarchical pyramid conditions with the reset gate, our model can adaptively selects geometric features for denoising. The process of encoding the condition $c$ into the denoising framework is incorporated into the training objective in Eq.\ref{loss}:
\begin{equation}
\begin{split}
&\mathcal{L}_{dif}(\theta, \rho, \omega)= \\
&\mathbb{E}_{t, P^0, \epsilon}\left\|\epsilon-\epsilon_\theta\left(\sqrt{\bar{a}_t} P^0+\sqrt{1-\bar{a}_t} \epsilon, f_\omega\left(\Phi_\rho\right), t\right)\right\|^2, 
\end{split}
\end{equation}
where $f_\omega, \epsilon_\theta, \Phi_\rho$ represents the H2 Net, the DIP Net, and the encoder that needs to be trained respectively.

Intuitively, DIP Net enables the encoder to extract high-level geometric features from the original point cloud and allows noisy point clouds by leveraging hierarchical geometric features from different frequency domains. Ablation studies (see \textit{Section~\ref{abla}}) demonstrate the effectiveness of the dual-channel design, showing that the model can learn diverse geometric priors, leading to improved performance in downstream tasks.
\subsection{Diffusion Guided Contrastive Learning}
After the diffusion model completes its foundational role, our cross-modal contrastive learning model takes up the baton, seamlessly continuing the representing learning process to enhance feature alignment and data diversity. 

Given the sample data $D=$ $\left\{\left(P_i, I_i, T_i\right)\right\}_{i=1}^{|D|}$, where $P_i, I_i, T_i$ denote 3D point cloud, 2D image, and text. A point cloud decoder is then employed to extract global feature $H_{\mathrm{pc}}$. For 2D images and text, we adopt a similar approach to embed them into a feature space using the frozen backbone Vision Transformer (VIT-B) and CLIP, obtaining $H_{\text {img }}$ and $H_{\text {text }}$. To further enhance the discriminative capability of the point cloud model, features from text and images are used to maximize the similarity between $H_{\text {pc }}$ and themselves. 

We compute the loss function $l\left(i,z,h\right)$ for the positive pair of examples ${z}_i$ and ${h}_i$ as:
\begin{equation}
\begin{split}
&l(i, z, h) \\
&=-\log \frac{\exp \left(s\left(z_i, h_i\right) / \tau\right)}{\sum_{\substack{k=1 \\ k \neq i}}^N \exp \left(s\left(z_i, z_k\right) / \tau\right)+\sum_{k=1}^N \exp \left(s\left(z_i, h_k\right) / \tau\right)},
\end{split}
\end{equation}
where N is the mini-batch size, $\tau$ is the temperature co-efficient and $s($.$)$ denotes the cosine similarity function. The cross-modal loss function for a mini-batch is then formulated as: 
\begin{equation}
\mathcal{L}_{}=\frac{1}{2 N} \sum_{i=1}^N[l(i, z, h)+l(i, h, z)].
\end{equation}

The overall contrastive loss function is composed of two components:
\begin{equation}
\mathcal{L}_{\text {con }}=\mathcal{L}\left(H_{\mathrm{pc}}, H_{\mathrm{img}}\right)+\mathcal{L}\left(H_{\mathrm{pc}}, H_{\text {text }}\right) .
\end{equation}

To prevent conflicts between diffusion and contrast, we apply a stop-gradient mechanism to all cross-attention modules, ensuring that the contrastive model does not propagate noisy gradients or training signals back to the diffusion model. Ablation studies (see \textit{Section~\ref{abla}}) demonstrate the effectiveness of this stop-gradient mechanism.

Finally, we derive the overall objective, which integrates the reconstruction loss from the diffusion model and the cross-modal contrastive loss. In this way, our PointDico is trained with merits from both contrastive and denoising aspects:
\begin{equation}
\mathcal{L}_{\mathrm{total}}=\mathcal{L}_{\mathrm{dif}}+\mathcal{L}_{\mathrm{con}}.
\end{equation}

\begin{figure}[t]
\centering
\includegraphics[width=1.0\columnwidth]{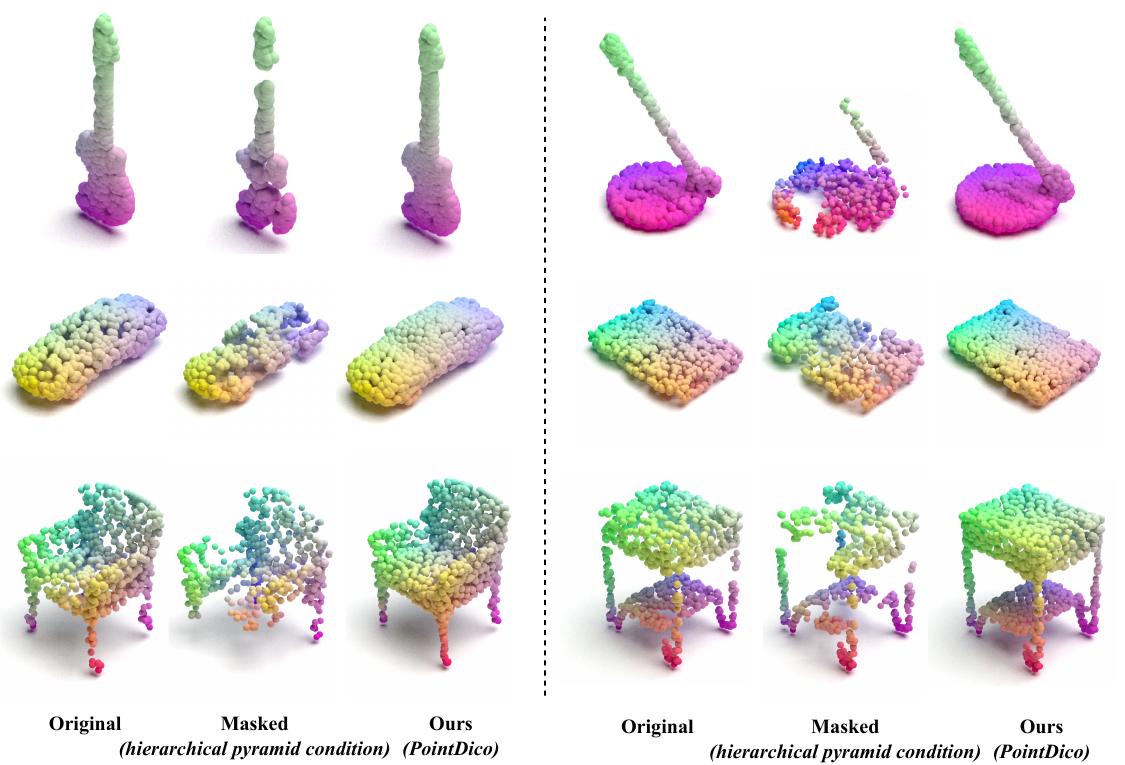}
\caption{\textbf{Visualization results on the ShapeNet dataset.} Each row visualizes the input point cloud, masked point cloud, and reconstructed point cloud. Even though we mask 60\% points, PointDico still produces high-quality point clouds.}
\label{vis}
\end{figure}

\section{Experiments}
\subsection{Pretraining}
\label{encoder}
\textbf{Setups.} Our model is pretrained on ShapeNet~\cite{chang2015shapenet}, which contains 55 categories and over 50,000 CAD models. For each point cloud in the dataset, we utilize high-quality rendered images generated by \cite{qi2023contrast}, along with category labels and manually crafted prompt templates. This method leverages the large-scale nature of RGB images and free-form language to significantly enrich the limited 3D data. Each point cloud in ShapeNet contains 2,048 points with only $x, y, z$ coordinates. All rendered images are resized to $224 \times 224$ pixels, with a patch size of 16 to fit the image encoder. 

\textbf{Encoders.} A mini-PointNet~\cite{qi2017pointnet} is employed for 3D patch embedding. We use Vision Transformer (ViT-B)~\cite{dosovitskiy2020image} pretrained on ImageNet as the image encoder and CLIP~\cite{radford2021learning} as the text encoder. For denoising generative modeling, we use ``FPS+KNN" to divide the input point cloud into 64 point patches, each containing 32 points. For contrastive learning, the image and text encoders are frozen, and Smooth $\ell_1$ loss is used. We also analyze the pretraining targets and masking ratio, with further details provided in Table \ref{bp} and Fig.\ref{mask}.

\textbf{Hyperparameters.} We conduct pretraining in 300 epochs with a learning rate of $5 \times 10^{-4}$. We use AdamW as the optimizer, a batch size of 64 , and a drop path rate of 0.1. All pretraining and downstream task experiments are conducted on a single NVIDIA RTX 4090.

\textbf{Visualization.} To demonstrate the effectiveness of our pretraining scheme, we provide a series of visualization generated by \textit{PointDico}. As shown in Fig.\ref{vis}., a mask ratio of 0.6 is applied to the input point cloud for masking, and the masked point cloud is used as a hierarchical pyramid condition to guide the diffusion model in generating the original point cloud. Our PointDico is capable of generating high-quality point clouds. Experimental results demonstrate the efficacy of the geometric prior learned through our pretraining framework, which provides robust guidance for shallow texture and shape semantics.
\subsection{Transfer Learning on Downstream Tasks}
We evaluated the transferability and effectiveness of our \textit{PointDico} on four widely used downstream tasks in point cloud representation learning: (I) 3D Real-World Object Classification, (II) 3D Synthetic Object Classification,  (III) 3D Zero-Shot Recognition. (IV) 3D Part Segmentation.

\textbf{3D Real-World Object Classification.} 
ScanObjectNN~\cite{uy2019revisiting} is one of the most challenging real-world datasets in the 3D domain, encompassing 15 categories and approximately 15,000 real-world objects. To ensure a fair comparison with existing methods, we present the experimental results both with and without the application of a voting strategy. Table \ref{cls} reveals that on the three ScanObjectNN benchmark variants, our method demonstrates substantial performance improvements of \textbf{+12.53\%}, \textbf{+10.15\%}, and \textbf{+3.21\%} respectively, compared to the three baseline models. Notably, in OBJ\_BG benchmark, our model achieves an accuracy of 94.32\% with the voting strategy, exhibiting significant improvements over the MAE-based baseline (+4.30\%), the cross-modal baseline (+3.38\%), and the Diffusion-based baseline (+1.03\%).

\begin{table}[t]
\centering
\caption{\textbf{Classification results on the ScanObjectNN and ModelNet40 datasets.}  Overall accuracy (\%) is reported for both without voting and with voting.} 
\resizebox{\linewidth}{!}{
\begin{tabular}{lccccc}
\hline
\multirow{2}{*}{Method}                                                                   & \multicolumn{3}{c}{ScanObjectNN}                 & \multicolumn{2}{c}{ModelNet40} \\ \cline{2-6} 
                                                                                          & OBJ\_BG        & OBJ\_ONLY      & PB\_T50\_RS    & 1k P           & 8k P          \\ \hline
\multicolumn{6}{c}{Supervised Learning Only}                                                                                                                                  \\ \hline
PointNet~\cite{qi2017pointnet}                                    & 73.3           & 79.2           & 68.0           & 89.2           & 90.8          \\
PointNet++~\cite{qi2017pointnet++}                                  & 82.3           & 84.3           & 77.9           & 90.7           & 91.9          \\
DGCNN~\cite{wang2019dynamic}                                      & 82.8           & 86.2           & 78.1           & 92.9           & -             \\
PointCNN~\cite{NEURIPS2018_f5f8590c}                            & 86.1           & 85.5           & 78.5           & 92.2           & -             \\
SimpleView~\cite{Goyal2021RevisitingPC}                     & -              & -              & 80.5           & 93.9           & -             \\
PCT~\cite{Guo2020PCTPC}                                         & -              & -              & -              & 93.2           & -             \\
PointMLP~\cite{ma2022rethinking}                                 & -              & -              & 85.4           & 94.5           & -             \\
PointNeXt~\cite{qian2022pointnext}                                  & -              & -              & 87.7           & 94.0           & -             \\
P2P-RN101~\cite{wang2022p2p}                                     & -              & -              & 87.4           & 93.1           & -             \\
P2P-HorNet~\cite{wang2022p2p}                                      & -              & -              & 89.3           & 94.0           & -             \\ \hline
\multicolumn{6}{c}{\textit{with Self-Supervised Representation Learning (MAE-based)}}                                                                                         \\ \hline
Point-BERT~\cite{yu2022point}                                      & 87.43          & 88.12          & 83.07          & 93.2           & 93.8          \\
MaskPoint~\cite{liu2022masked}                                    & 89.30          & 88.10          & 84.30          & 93.8           & -             \\
\rowcolor[HTML]{F0ECF7}
Point-MAE~\cite{pang2022masked}                                   & 90.02          & 88.29          & 85.18          & 93.8           & 94.0          \\
Point-M2AE~\cite{zhang2022point}                                   & 91.22          & 88.81          & 86.43          & 94.0           & -             \\
\rowcolor[HTML]{cef6eb} 
\textit{Improvement(baseline:PointMAE)}                                                   & \textbf{+4.30} & \textbf{+4.31} & \textbf{+3.92} & \textbf{-}     & \textbf{-}    \\ \hline
\multicolumn{6}{c}{\textit{with Cross-Modal Self-Supervised Learning}}                                                                                                        \\ \hline
\rowcolor[HTML]{F0ECF7}
Joint-MAE~\cite{guo2023joint}                                      & 90.94          & 88.86          & 86.07          & 94.0           & -             \\
I2P-MAE~\cite{zhang2022learning3drepresentations2d}              & 94.15          & 91.57          & 90.11          & 93.7           & -             \\
ACT~\cite{dong2023autoencoders}                                    & 93.29          & 91.91          & 88.21          & 93.7           & 94.0          \\
\rowcolor[HTML]{cef6eb} 
\textit{Improvement(baseline:Joint-MAE)}                                                  & \textbf{+3.38} & \textbf{+3.74} & \textbf{+3.03} & \textbf{-}     & -             \\ \hline
\multicolumn{6}{c}{\textit{with Self-Supervised Representation Learning (Diffusion-based)}}                                                                                   \\ \hline
\rowcolor[HTML]{F0ECF7}
PointDif~\cite{zheng2024point}                                                                                  & 93.29          & 91.91          & 87.61          & 91.8           & 91.9          \\
\rowcolor[HTML]{ffe6df} 
\textbf{PointDico} w/o vot. & \textbf{93.80} & \textbf{92.08} & \textbf{88.51} & \textbf{92.2}  & \textbf{92.5} \\
\rowcolor[HTML]{ffe6df}  
\textbf{PointDico} w/ vot.  & \textbf{94.32} & \textbf{92.60} & \textbf{89.10} & \textbf{92.4}  & \textbf{92.6} \\
\rowcolor[HTML]{cef6eb} 
\textit{Improvement(baseline:PointDif)}                                                   & \textbf{+1.03} & \textbf{+0.69} & \textbf{+1.49} & \textbf{+0.6}  & \textbf{+0.7} \\ \hline
\label{cls}
\end{tabular}}
\end{table}

\textbf{3D Synthetic Object Classification}. The ModelNet dataset~\cite{wu20153d}, introduced by Wu et al. in 2015, serves as a foundational benchmark for synthetic 3D object recognition. It consists of 12,000 meshed 3D CAD models, categorized into 40 classes according to the ModelNet40 configuration. During fine-tuning, we adhere to the procedures outlined in PointNet~\cite{qi2017pointnet}, applying scaling and translation operations for data augmentation. The results, summarized in Table \ref{cls}, indicate that our PointDico achieves an accuracy of 92.2\% without voting and 92.4\% with voting. Although synthetic datasets cannot fully simulate the complexity of real-world objects, our PointDico still outperforms other diffusion-based model, with an overall improvement of \textbf{+1.3\%}. This suggests that our model exhibits superior generalization performance, as it is capable of extracting efficient representations even in datasets with limited noise and occlusion.

\textbf{3D Part Segmentation.} To evaluate geometric understanding at the object scale, we performed part segmentation experiments on the ShapeNetPart dataset~\cite{10.1145/2980179.2980238}. Specifically, we integrated cross-modal features into global representations and utilized the same segmentation head as employed in PointMAE ~\cite{pang2022masked}. As shown in Table \ref{seg}, our model surpasses the baseline by \textbf{+1.0\%} in Cls.mIoU and \textbf{+0.9\%} in Inst.mIoU. This improvement can be attributed to the superior geometric representation capabilities of the diffusion method, allowing PointDico to achieve enhanced shape understanding.

\begin{table}[t]
\centering
\caption{\textbf{Part segmentation on ShapeNetPart dataset.} The mIoU(\%) overall classes (Cls.) and the mIoU(\%) overall instances (Inst.) are reported. The dagger symbol ${ } ^ {\dagger}$ indicates that the model was reproduced.}
\resizebox{\linewidth}{!}{
\begin{tabular}{lcc}
\hline
\textbf{Methods}                   & \textbf{Cls. mIoU (\%)} & \textbf{Inst. mIoU (\%)} \\ \hline
PointNet~\cite{qi2017pointnet}        & 80.4                    & 83.7                     \\
PointNet++~\cite{qi2017pointnet++}     & 81.9                    & 85.1                     \\
DGCNN~\cite{wang2019dynamic}         & 82.3                    & 85.2                     \\
PointMLP~\cite{ma2022rethinking}      & 84.6                    & 86.1                     \\ \hline
Transformer~\cite{vaswani2017attention} & 83.6                    & 85.2                     \\
PointContrast~\cite{xie2020pointcontrastunsupervisedpretraining3d} & -                       & 85.1                     \\
CrossPoint~\cite{afham2022crosspoint}   & -                       & 85.5                     \\
Point-BERT~\cite{yu2022point}     & 84.1                    & 85.6                     \\
ACT~\cite{dong2023autoencoders}    & 84.7                    & 86.1                     \\
Point-MAE~\cite{pang2022masked}      & 84.4                    & 86.1                     \\
EDGCNet~\cite{cheng_edgcnet_2024}&84.4&86.7\\
\rowcolor[HTML]{F0ECF7}
PointDif${ } ^ {\dagger}$~\cite{zheng2024point}                              & 83.5                    & 85.6                     \\
\rowcolor[HTML]{ffe6df} 
\textbf{PointDico}                         & \textbf{84.5}           & \textbf{86.5}          \\ 
\rowcolor[HTML]{cef6eb} 
\textit{Improvement(baseline:PointDif)}                                                   & \textbf{+1.0} & \textbf{+0.9}   \\ \hline
\end{tabular}}
\label{seg}
\end{table}

\begin{table}[t]
\caption{Zero-shot 3D object classification on ModelNet40
 (MN-40) and ModelNet10(MN-10). Top-1 accuracy(\%) is reported. Ensemb. denotes whether to use the ensemble strategy
 with multiple text inputs.}
\resizebox{\linewidth}{!}{
\begin{tabular}{lcccc}
\hline
\textbf{Method}                 & \textbf{Backbone} & \textbf{Ensemb.} & \textbf{MN-10} & \textbf{MN-40} \\ \hline
PointCLIP~\cite{zhang2022pointclip} & ResNet-50         & $\times$         & 30.2           & 20.2           \\
CLIP2Point~\cite{huang2023clip2point} & Transformer       & $\checkmark$     & 66.6           & 49.4           \\
\textbf{PointDico }                      & Transformer       & $\times$         & \textbf{70.4}         & \textbf{56.4}          \\
\rowcolor[HTML]{ffe6df} 
\textbf{PointDico}                       & Transformer       & $\checkmark$     & \textbf{75.7}        & \textbf{60.6}           \\ \hline
\end{tabular}}
\label{zero}
\end{table}

\textbf{3D Zero-Shot Recognition.} Our model aligns the feature spaces of text and other modalities. Consequently, our model exhibits a more efficient ability to represent data. We utilize the ModelNet40 dataset~\cite{wu20153d} for zero-shot evaluation. Following PointCLIP~\cite{zhang2022pointclip}, we employ prompt templates with the category label to generate text features. As shown in Table \ref{zero}, our PointDico surpasses all zero-shot methods with CNN-based or Transformer-based backbones. Furthermore, by employing ensemble methods~\cite{huang2023clip2point} such as multi-prompt templates, our PointDico achieves a Top-1 accuracy of \textbf{60.6\%} on ModelNet 40, significantly outperforming PointCLIP and CLIP2Point by \textbf{40.4\%} and \textbf{11.2\%}, respectively.

\subsection{Ablations and Analysis}
\label{abla}
\begin{table}[t]
\caption{\textbf{Ablation study on pretraining targets.} The table reports the overall accuracy (\%) w/ voting on the ScanObjectNN
OBJ\_BG and ModelNet40 dataset.}
\centering
\resizebox{0.8\linewidth}{!}{
\begin{tabular}{cccccc}
\hline
\multirow{2}{*}{\textbf{Diffusion}} & \multicolumn{3}{c}{\textbf{Contrastive}}            & \multirow{2}{*}{\begin{tabular}[c]{@{}c@{}}\textbf{ScanObjectNN}\\ OBJ\_BG\end{tabular}} & \multirow{2}{*}{\textbf{MN40}} \\ \cline{2-4}
                           & text         & img          & self         &                                                                                 &                       \\ \hline
                           \rowcolor[HTML]{F0ECF7}
$\times$                   & $\times$     & $\times$     & $\checkmark$ & 90.02                                                                           & 89.2                  \\
$\times$                   & $\checkmark$ & $\checkmark$ & $\times$     & 90.11                                                                           & 90.9                  \\
$\times$                   & $\checkmark$ & $\checkmark$ & $\checkmark$ & 91.22                                                                           & 90.8                  \\ \hline
\rowcolor[HTML]{F0ECF7}
$\checkmark$               & $\times$     & $\times$  & $\times$  & 92.60                                                                           & 91.2                  \\
$\checkmark$               & $\checkmark$ & $\times$     & $\checkmark$ & 93.29                                                                           & 91.6                  \\
\rowcolor[HTML]{ffe6df} 
$\checkmark$               & $\checkmark$ & $\checkmark$ & $\times$     & \textbf{94.32}                                                                           & \textbf{92.4}                 \\
$\checkmark$               & $\checkmark$ & $\checkmark$ & $\checkmark$ & 94.15                                                                           & 92.2                  \\ \hline
\end{tabular}}
\label{tar}
\end{table}

\begin{table}[t]
\caption{\textbf{Ablation study on conditional guidance strategies.} We report the Classes mIoU(\%) and Instances mIoU(\%) on ShapeNetPart}
\centering
\begin{tabular}{lcc}
\hline
\textbf{Methods}     & \textbf{Cls.mIoU(\%)} & \textbf{Inst.mIoU(\%)} \\ \hline
Cross Attention      & 82.3                  & 85.1                   \\
Point Concat         & 82.3                  & 85.2                   \\
\rowcolor[HTML]{F0ECF7}
PC Network~\cite{zheng2024point}           & 83.5                  & 85.6                   \\
\rowcolor[HTML]{ffe6df} 
\textbf{DIP Network} & \textbf{84.5}         & \textbf{86.5}          \\ \hline
\end{tabular}
\label{stra}
\end{table}

\textbf{Pretraining Targets.} To further analyze the significance of each module and validate the effectiveness of our diffusion-guided contrastive learning, we conducted ablation experiments on the model's primary targets. The results are presented in Table \ref{tar}. When the diffusion-guided approach is excluded, the model performs poorly due to the limited three-dimensional data available for fitting. Additionally, the performance of the single-modal contrastive learning is inferior to that of the multi-modal approach, which corroborates our hypothesis that multi-modal data enhances the model's generalization ability. It is noteworthy that introducing self-modal contrastive learning actually leads to a decrease in accuracy. We attribute this decline to the excessive number of loss functions, which may compete with each other, making the optimization goals unclear and complicating the training process.

\textbf{Conditional Guidance Strategies.} We investigated the impact of various guidance strategies on the ShapeNetPart~\cite{10.1145/2980179.2980238} dataset. As shown in Table \ref{stra}, DIP Net achieved the state-of-the-art performance. Cross-attention focuses on static pairwise relationships, which may fail to capture complex interactions in 3D data. In contrast, DIP Net improves adaptability and accuracy by dynamically adjusting feature representations at each layer. Additionally, compared to PCNet~\cite{zheng2024point}, DIP Net refines features progressively across multiple scales, capturing fine-grained details at lower scales and global semantic information at higher scales, thus enhancing performance.

\begin{figure}[t]
\centering
\hspace{-0.5cm}
\includegraphics[scale=0.3]{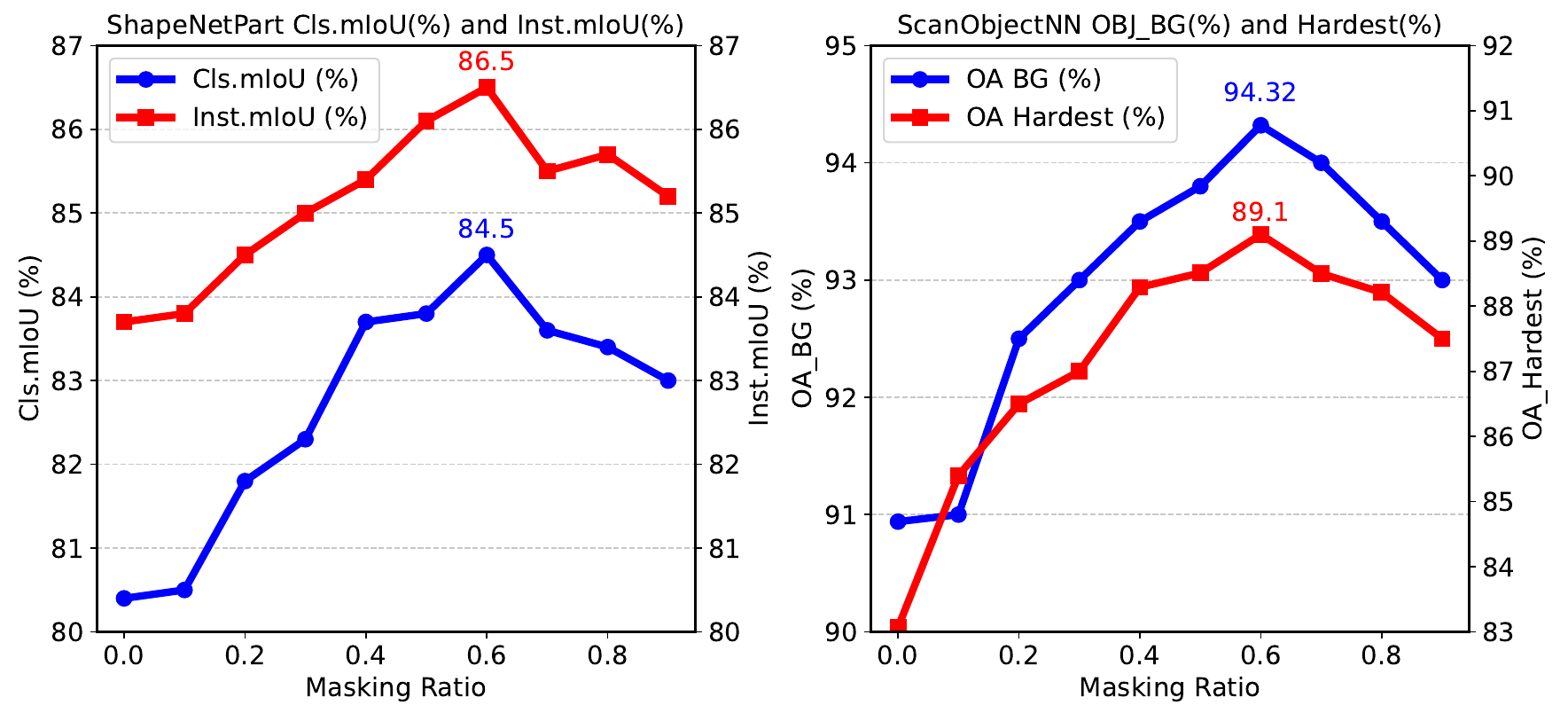}
\caption{\textbf{Ablation study on masking ratio.} We report the Overall Accuracy(\%) on ScanObjectNN and the mIoU(\%) on ShapeNetPart with different
 masking ratios.}
\label{mask}
\end{figure}

\textbf{Masking Ratio.} We examine the effect of different masking ratios on downstream tasks, reporting results for classification on ScanObjectNN~\cite{uy2019revisiting} and part segmentation on ShapeNetPart~\cite{10.1145/2980179.2980238}. As shown in Fig \ref{mask}, encoding all point patches without masking harms the model's learning. Masking increases the difficulty of the self-supervised proxy task, helping the backbone learn richer geometric priors. Moreover, our method achieves optimal classification and part segmentation performance at the mask ratio of 0.6.

\begin{table}[t]
\caption{\textbf{Ablation study on back-propagation operations and backbone.} The table reports the overall accuracy (\%) with voting on the ScanObjectNN OBJ\_BG dataset.}
\begin{tabular}{cccc}
\hline
\multirow{2}{*}{\textbf{Backbone}} & \multicolumn{2}{c}{\textbf{Back-propagation}} & \multirow{2}{*}{\textbf{Accuracy}} \\ \cline{2-3}
                                   & 3D Encoder         & 2D \& Text Encoder       &                                    \\ \hline
ViT-B \& CLIP                      & $\checkmark$       & $\checkmark$             & 90.02                              \\
\rowcolor[HTML]{ffe6df} 
ViT-B \& CLIP                      & $\times$           & $\times$                 & \textbf{94.32}                     \\
ViT-B \& CLIP                      & $\checkmark$       & $\times$                 & 93.29                              \\ \hline
Resnet34 \& CLIP                   & $\checkmark$       & $\checkmark$             & 87.13                              \\
Resnet34 \& CLIP                   & $\times$           & $\times$                 & 94.15                              \\
Resnet34 \& CLIP                   & $\checkmark$       & $\times$                 & 93.29                              \\ \hline
\end{tabular}
\label{bp}
\end{table}

\textbf{Encoder with Back-Propagation Operation.} In addition to the aforementioned experiments, we explored strategies for identifying better 2D and text encoder teachers to further enhance model performance. The results, summarized in Table \ref{bp}, show that using a Vision Transformer (ViT-B) as the 2D encoder teacher outperforms architectures based on ResNet. Furthermore, we found that applying back-propagation directly on the 3D encoder reduces model performance. We argue that this decline is primarily due to the limited 3D data sending noisy signals to the diffusion model during pretraining, which interferes with the guidance. Additionally, unfreezing the 2D and text encoders leads to a significant increase in the number of model parameters, which not only increases computational overhead but also prevents the model from learning effective 3D representations.

\section{CONCLUSION}
In this paper, we propose a novel diffusion-guided contrastive learning framework for point cloud pretraining, called PointDico. It leverages a hierarchical pyramid conditional generator to assist the point cloud backbone in learning geometric priors. H2 Net excels at efficiently capturing multi-scale features, while adaptive conditional fusion mechanism in DIP Net enhances model flexibility. Our approach integrates diffusion models with contrastive learning. Compared to existing methods, PointDico more effectively utilizes multi-modal data while preserving hierarchical semantic information. Overall, our diffusion-guided contrastive pretraining framework offers a new direction for advancing point cloud processing.

\section*{Acknowledgment}
This work is supported by the Fundamental Research Funds for Central Universities under grant No. xzy012024105.

\bibliography{ijcnn2025}

@inproceedings{Goyal2021RevisitingPC,
  title={Revisiting Point Cloud Shape Classification with a Simple and Effective Baseline},
  author={Ankit Goyal and Hei Law and Bowei Liu and Alejandro Newell and Jia Deng},
  booktitle={International Conference on Machine Learning},
  year={2021},
  url={https://api.semanticscholar.org/CorpusID:235390998}
}

@article{10.1145/2980179.2980238,
author = {Yi, Li and Kim, Vladimir G. and Ceylan, Duygu and Shen, I-Chao and Yan, Mengyan and Su, Hao and Lu, Cewu and Huang, Qixing and Sheffer, Alla and Guibas, Leonidas},
title = {A scalable active framework for region annotation in 3D shape collections},
year = {2016},
issue_date = {November 2016},
publisher = {Association for Computing Machinery},
address = {New York, NY, USA},
volume = {35},
number = {6},
issn = {0730-0301},
url = {https://doi.org/10.1145/2980179.2980238},
doi = {10.1145/2980179.2980238},
journal = {ACM Trans. Graph.},
month = dec,
articleno = {210},
numpages = {12}
}

@misc{xie2020pointcontrastunsupervisedpretraining3d,
      title={PointContrast: Unsupervised Pre-training for 3D Point Cloud Understanding}, 
      author={Saining Xie and Jiatao Gu and Demi Guo and Charles R. Qi and Leonidas J. Guibas and Or Litany},
      year={2020},
      eprint={2007.10985},
      archivePrefix={arXiv},
      primaryClass={cs.CV},
      url={https://arxiv.org/abs/2007.10985}, 
}

@article{zha2023towards,
  title={Towards Compact 3D Representations via Point Feature Enhancement Masked Autoencoders},
  author={Zha, Yaohua and Ji, Huizhen and Li, Jinmin and Li, Rongsheng and Dai, Tao and Chen, Bin and Wang, Zhi and Xia, Shu-Tao},
  journal={arXiv preprint arXiv:2312.10726},
  year={2023}
}

@misc{zhang2022learning3drepresentations2d,
      title={Learning 3D Representations from 2D Pre-trained Models via Image-to-Point Masked Autoencoders}, 
      author={Renrui Zhang and Liuhui Wang and Yu Qiao and Peng Gao and Hongsheng Li},
      year={2022},
      eprint={2212.06785},
      archivePrefix={arXiv},
      primaryClass={cs.CV},
      url={https://arxiv.org/abs/2212.06785}, 
}

@article{cheng_edgcnet_2024,
	title = {{EDGCNet}: {Joint} dynamic hyperbolic graph convolution and dual squeeze-and-attention for {3D} point cloud segmentation},
	volume = {237},
	issn = {0957-4174},
	url = {https://www.sciencedirect.com/science/article/pii/S0957417423020535},
	doi = {https://doi.org/10.1016/j.eswa.2023.121551},
	journal = {Expert Systems with Applications},
	author = {Cheng, Haozhe and Zhu, Jihua and Lu, Jian and Han, Xu},
	year = {2024},
	pages = {121551},
}

@article{Guo2020PCTPC,
  title={PCT: Point cloud transformer},
  author={Meng-Hao Guo and Junxiong Cai and Zheng-Ning Liu and Tai-Jiang Mu and Ralph Robert Martin and Shimin Hu},
  journal={Computational Visual Media},
  year={2020},
  volume={7},
  pages={187 - 199}
}

@misc{zhang2024pcpmaelearningpredictcenters,
      title={PCP-MAE: Learning to Predict Centers for Point Masked Autoencoders}, 
      author={Xiangdong Zhang and Shaofeng Zhang and Junchi Yan},
      year={2024},
      eprint={2408.08753},
      archivePrefix={arXiv},
      primaryClass={cs.CV},
      url={https://arxiv.org/abs/2408.08753}, 
}

@inproceedings{luo2021diffusion,
  author = {Luo, Shitong and Hu, Wei},
  title = {Diffusion Probabilistic Models for 3D Point Cloud Generation},
  booktitle = {Proceedings of the IEEE/CVF Conference on Computer Vision and Pattern Recognition (CVPR)},
  month = {June},
  year = {2021}
}

@misc{grill2020bootstraplatentnewapproach,
      title={Bootstrap your own latent: A new approach to self-supervised Learning}, 
      author={Jean-Bastien Grill and Florian Strub and Florent Altché and Corentin Tallec and Pierre H. Richemond and Elena Buchatskaya and Carl Doersch and Bernardo Avila Pires and Zhaohan Daniel Guo and Mohammad Gheshlaghi Azar and Bilal Piot and Koray Kavukcuoglu and Rémi Munos and Michal Valko},
      year={2020},
      eprint={2006.07733},
      archivePrefix={arXiv},
      primaryClass={cs.LG},
      url={https://arxiv.org/abs/2006.07733}, 
}

@article{Ramesh2022HierarchicalTI,
  title={Hierarchical Text-Conditional Image Generation with CLIP Latents},
  author={Aditya Ramesh and Prafulla Dhariwal and Alex Nichol and Casey Chu and Mark Chen},
  journal={ArXiv},
  year={2022},
  volume={abs/2204.06125},
  url={https://api.semanticscholar.org/CorpusID:248097655}
}

@inproceedings{wei2023diffusion,
      author    = {Wei, Chen and Mangalam, Karttikeya and Huang, Po-Yao and Li, Yanghao and Fan, Haoqi and Xu, Hu and Wang, Huiyu and Xie, Cihang and Yuille, Alan and Feichtenhofer, Christoph},
      title     = {Diffusion Models as Masked Autoencoder},
      booktitle = {ICCV},
      year      = {2023},
    }

@inproceedings{10.5555/3600270.3600898,
author = {Ho, Jonathan and Salimans, Tim and Gritsenko, Alexey and Chan, William and Norouzi, Mohammad and Fleet, David J.},
title = {Video diffusion models},
year = {2024},
isbn = {9781713871088},
publisher = {Curran Associates Inc.},
address = {Red Hook, NY, USA},
booktitle = {Proceedings of the 36th International Conference on Neural Information Processing Systems},
articleno = {628},
numpages = {14},
location = {New Orleans, LA, USA},
series = {NIPS '22}
}

@InProceedings{pmlr-v37-sohl-dickstein15,
  title = 	 {Deep Unsupervised Learning using Nonequilibrium Thermodynamics},
  author = 	 {Sohl-Dickstein, Jascha and Weiss, Eric and Maheswaranathan, Niru and Ganguli, Surya},
  booktitle = 	 {Proceedings of the 32nd International Conference on Machine Learning},
  pages = 	 {2256--2265},
  year = 	 {2015},
  editor = 	 {Bach, Francis and Blei, David},
  volume = 	 {37},
  series = 	 {Proceedings of Machine Learning Research},
  address = 	 {Lille, France},
  month = 	 {07--09 Jul},
  publisher =    {PMLR},
  url = 	 {https://proceedings.mlr.press/v37/sohl-dickstein15.html}
}

@inproceedings{NEURIPS2018_f5f8590c,
 author = {Li, Yangyan and Bu, Rui and Sun, Mingchao and Wu, Wei and Di, Xinhan and Chen, Baoquan},
 booktitle = {Advances in Neural Information Processing Systems},
 editor = {S. Bengio and H. Wallach and H. Larochelle and K. Grauman and N. Cesa-Bianchi and R. Garnett},
 pages = {},
 publisher = {Curran Associates, Inc.},
 title = {PointCNN: Convolution On X-Transformed Points},
 volume = {31},
 year = {2018}
}

@inproceedings{devlin2018bert,
  author       = {Devlin, Jacob and
                  Chang, MingWei and
                  Lee, Kenton and
                  Toutanova, Kristina},
  title        = {{BERT:} Pre-training of Deep Bidirectional Transformers for Language
                  Understanding},
  booktitle    = {Proceedings of the 2019 Conference of the North American Chapter of
                  the Association for Computational Linguistics: Human Language Technologies},
  pages        = {4171--4186},
  publisher    = {Association for Computational Linguistics},
  year         = {2019},
}

@inproceedings{chen2020simple,
  title={A simple framework for contrastive learning of visual representations},
  author={Chen, Ting and Kornblith, Simon and Norouzi, Mohammad and Hinton, Geoffrey},
  booktitle={Proceedings of International Conference on Machine Learning (ICML)},
  pages={1597--1607},
  year={2020}
}

@inproceedings{he2022masked,
  title={Masked autoencoders are scalable vision learners},
  author={He, Kaiming and Chen, Xinlei and Xie, Saining and Li, Yanghao and Doll{\'a}r, Piotr and Girshick, Ross},
  booktitle={Proceedings of the IEEE/CVF Conference on Computer Vision and Pattern Recognition (CVPR)},
  pages={16000--16009},
  year={2022}
}

@inproceedings{afham2022crosspoint,
  title={Crosspoint: Self-supervised cross-modal contrastive learning for 3d point cloud understanding},
  author={Afham, Mohamed and Dissanayake, Isuru and Dissanayake, Dinithi and Dharmasiri, Amaya and Thilakarathna, Kanchana and Rodrigo, Ranga},
  booktitle={Proceedings of the IEEE/CVF Conference on Computer Vision and Pattern Recognition (CVPR)},
  pages={9902--9912},
  year={2022}
}

@inproceedings{pang2022masked,
  title={Masked autoencoders for point cloud self-supervised learning},
  author={Pang, Yatian and Wang, Wenxiao and Tay, Francis EH and Liu, Wei and Tian, Yonghong and Yuan, Li},
  booktitle={Proceedings of European Conference on Computer Vision (ECCV)},
  pages={604--621},
  year={2022},
  organization={Springer}
}

@inproceedings{zhang2022point,
  title={Point-m2ae: multi-scale masked autoencoders for hierarchical point cloud pre-training},
  author={Zhang, Renrui and Guo, Ziyu and Gao, Peng and Fang, Rongyao and Zhao, Bin and Wang, Dong and Qiao, Yu and Li, Hongsheng},
  booktitle={Proceedings of Advances in Neural Information Processing Systems (NeurIPS)},
  year={2022}
}

@inproceedings{liu2022masked,
  title={Masked discrimination for self-supervised learning on point clouds},
  author={Liu, Haotian and Cai, Mu and Lee, Yong Jae},
  booktitle={Proceedings of European Conference on Computer Vision (ECCV)},
  pages={657--675},
  year={2022},
  organization={Springer}
}

@inproceedings{dong2023autoencoders,
  title={Autoencoders as Cross-Modal Teachers: Can Pretrained 2D Image Transformers Help 3D Representation Learning?},
  author={Dong, Runpei and Qi, Zekun and Zhang, Linfeng and Zhang, Junbo and Sun, Jianjian and Ge, Zheng and Yi, Li and Ma, Kaisheng},
  booktitle={Proceedings of International Conference on Learning Representations (ICLR)},
  year={2023}
}

@inproceedings{qi2023contrast,
  title={Contrast with Reconstruct: Contrastive 3D Representation Learning Guided by Generative Pretraining},
  author={Qi, Zekun and Dong, Runpei and Fan, Guofan and Ge, Zheng and Zhang, Xiangyu and Ma, Kaisheng and Yi, Li},
  booktitle={Proceedings of International Conference on Machine Learning (ICML)},
  year={2023}
}

@inproceedings{guo2023joint,
  title     = {Joint-MAE: 2D-3D Joint Masked Autoencoders for 3D Point Cloud Pre-training},
  author    = {Guo, Ziyu and Zhang, Renrui and Qiu, Longtian and Li, Xianzhi and Heng, Pheng-Ann},
  booktitle = {Proceedings of International Joint Conference on Artificial Intelligence (IJCAI)},
  pages     = {791--799},
  year      = {2023}
}

@inproceedings{yu2022point,
  title={Point-bert: Pre-training 3d point cloud transformers with masked point modeling},
  author={Yu, Xumin and Tang, Lulu and Rao, Yongming and Huang, Tiejun and Zhou, Jie and Lu, Jiwen},
  booktitle={Proceedings of the IEEE/CVF Conference on Computer Vision and Pattern Recognition (CVPR)},
  pages={19313--19322},
  year={2022}
}

@inproceedings{vaswani2017attention,
  title={Attention is all you need},
  author={Vaswani, Ashish and Shazeer, Noam and Parmar, Niki and Uszkoreit, Jakob and Jones, Llion and Gomez, Aidan N and Kaiser, {\L}ukasz and Polosukhin, Illia},
  booktitle={Proceedings of Advances in Neural Information Processing Systems (NeurIPS)},
  year={2017}
}

@inproceedings{zhang2022pointclip,
  title={Pointclip: Point cloud understanding by clip},
  author={Zhang, Renrui and Guo, Ziyu and Zhang, Wei and Li, Kunchang and Miao, Xupeng and Cui, Bin and Qiao, Yu and Gao, Peng and Li, Hongsheng},
  booktitle={Proceedings of the IEEE/CVF Conference on Computer Vision and Pattern Recognition (CVPR)},
  pages={8552--8562},
  year={2022}
}

@inproceedings{radford2021learning,
  title={Learning transferable visual models from natural language supervision},
  author={Radford, Alec and Kim, Jong Wook and Hallacy, Chris and Ramesh, Aditya and Goh, Gabriel and Agarwal, Sandhini and Sastry, Girish and Askell, Amanda and Mishkin, Pamela and Clark, Jack and others},
  booktitle={Proceedings of International Conference on Machine Learning (ICML)},
  pages={8748--8763},
  year={2021}
}

@article{chang2015shapenet,
  title={Shapenet: An information-rich 3d model repository},
  author={Chang, Angel X and Funkhouser, Thomas and Guibas, Leonidas and Hanrahan, Pat and Huang, Qixing and Li, Zimo and Savarese, Silvio and Savva, Manolis and Song, Shuran and Su, Hao and others},
  journal={arXiv preprint arXiv:1512.03012},
  year={2015}
}

@inproceedings{dosovitskiy2020image,
  title={An image is worth 16x16 words: Transformers for image recognition at scale},
  author={Dosovitskiy, Alexey and Beyer, Lucas and Kolesnikov, Alexander and Weissenborn, Dirk and Zhai, Xiaohua and Unterthiner, Thomas and Dehghani, Mostafa and Minderer, Matthias and Heigold, Georg and Gelly, Sylvain and others},
  booktitle={Proceedings of International Conference on Learning Representations (ICLR)},
  year={2021}
}

@inproceedings{uy2019revisiting,
  title={Revisiting point cloud classification: A new benchmark dataset and classification model on real-world data},
  author={Uy, Mikaela Angelina and Pham, Quang-Hieu and Hua, Binh-Son and Nguyen, Thanh and Yeung, Sai-Kit},
  booktitle={Proceedings of the IEEE/CVF International Conference on Computer Vision (ICCV)},
  pages={1588--1597},
  year={2019}
}

@inproceedings{qi2017pointnet,
  title={Pointnet: Deep learning on point sets for 3d classification and segmentation},
  author={Qi, Charles R and Su, Hao and Mo, Kaichun and Guibas, Leonidas J},
  booktitle={Proceedings of the IEEE Conference on Computer Vision and Pattern Recognition (CVPR)},
  pages={652--660},
  year={2017}
}

@inproceedings{qi2017pointnet++,
  title={Pointnet++: Deep hierarchical feature learning on point sets in a metric space},
  author={Qi, Charles Ruizhongtai and Yi, Li and Su, Hao and Guibas, Leonidas J},
  booktitle={Proceedings of Advances in Neural Information Processing Systems (NeurIPS)},
  year={2017}
}

@article{wang2019dynamic,
  title={Dynamic graph cnn for learning on point clouds},
  author={Wang, Yue and Sun, Yongbin and Liu, Ziwei and Sarma, Sanjay E and Bronstein, Michael M and Solomon, Justin M},
  journal={ACM Transactions on Graphics (TOG)},
  volume={38},
  number={5},
  pages={1--12},
  year={2019},
  publisher={ACM New York, NY, USA}
}

@inproceedings{ma2022rethinking,
  title={Rethinking network design and local geometry in point cloud: A simple residual MLP framework},
  author={Ma, Xu and Qin, Can and You, Haoxuan and Ran, Haoxi and Fu, Yun},
  booktitle={Proceedings of International Conference on Learning Representations (ICLR)},
  year={2022}
}

@inproceedings{qian2022pointnext,
  title={Pointnext: Revisiting pointnet++ with improved training and scaling strategies},
  author={Qian, Guocheng and Li, Yuchen and Peng, Houwen and Mai, Jinjie and Hammoud, Hasan and Elhoseiny, Mohamed and Ghanem, Bernard},
  booktitle={Proceedings of Advances in Neural Information Processing Systems (NeurIPS)},
  pages={23192--23204},
  year={2022}
}

@inproceedings{wang2022p2p,
  title={P2p: Tuning pre-trained image models for point cloud analysis with point-to-pixel prompting},
  author={Wang, Ziyi and Yu, Xumin and Rao, Yongming and Zhou, Jie and Lu, Jiwen},
  booktitle={Proceedings of Advances in Neural Information Processing Systems (NeurIPS)},
  pages={14388--14402},
  year={2022}
}

@inproceedings{wu20153d,
  title={3d shapenets: A deep representation for volumetric shapes},
  author={Wu, Zhirong and Song, Shuran and Khosla, Aditya and Yu, Fisher and Zhang, Linguang and Tang, Xiaoou and Xiao, Jianxiong},
  booktitle={Proceedings of the IEEE Conference on Computer Vision and Pattern Recognition (CVPR)},
  pages={1912--1920},
  year={2015}
}

@inproceedings{huang2023clip2point,
  title={Clip2point: Transfer clip to point cloud classification with image-depth pre-training},
  author={Huang, Tianyu and Dong, Bowen and Yang, Yunhan and Huang, Xiaoshui and Lau, Rynson WH and Ouyang, Wanli and Zuo, Wangmeng},
  booktitle={Proceedings of the IEEE/CVF International Conference on Computer Vision (ICCV)},
  pages={22157--22167},
  year={2023}
}

@inproceedings{zheng2024point,
  title={Point Cloud Pre-training with Diffusion Models},
  author={Zheng, Xiao and Huang, Xiaoshui and Mei, Guofeng and Hou, Yuenan and Lyu, Zhaoyang and Dai, Bo and Ouyang, Wanli and Gong, Yongshun},
  booktitle={Proceedings of the IEEE/CVF Conference on Computer Vision and Pattern Recognition},
  pages={22935--22945},
  year={2024}
}
\vspace{12pt}
\color{red}

\end{document}